\documentclass[runningheads]{llncs}
\RequirePackage{silence}  
\usepackage{graphicx}
\usepackage{comment}
\usepackage{amsmath,amssymb}
\usepackage{color}
\usepackage{url}
\usepackage{hyperref}
\usepackage{wrapfig}
\usepackage[misc]{ifsym}

\usepackage{booktabs}
\usepackage{cleveref}
\usepackage{bbm}
\def\eg{\emph{e.g.}} 
\def\ie{\emph{i.e.}}

\newcommand{\methodname}{\emph{TinyVICL}}
\usepackage{fancyhdr}
\fancypagestyle{preprint}{\fancyhf{}\fancyfoot[L]{Preprint.}}

\usepackage{multirow}
\usepackage{array}

\newif\ifreview
\reviewtrue
\reviewfalse

\ifreview
	\usepackage{lineno}

	\linenumbers
\fi

\begin{document}


\def\SubNumber{18}

\def\GCPRTrack{Young Researcher's Forum}

\title{Beyond Model Size: Probing the Gaps in Visual in-Context Learning by Training a Tiny Model}

\ifreview
	\titlerunning{GCPR 2026 Submission \SubNumber{}. CONFIDENTIAL REVIEW COPY.}
	\authorrunning{GCPR 2026 Submission \SubNumber{}. CONFIDENTIAL REVIEW COPY.}
	\author{GCPR 2026 - \GCPRTrack{}}
	\institute{Paper ID \SubNumber}
\else

	\author{Sunil Khatri \and
	Steven Landgraf \and
	Markus Ulrich \and
	Simon Rei{\ss}$^{~\text{\Letter}}$}
	
	\authorrunning{S. Khatri et al.}
	
	\institute{Karlsruhe Institute of Technology \\
	\text{\Letter}~\email{simon.reiss@kit.edu}\\
    }
\fi

\maketitle              
\begin{abstract}
    Visual in-Context Learning (VICL) aims at making progress towards adaptive vision models, that can -- based on a few examples -- adapt to a new task at test-time.
    With the history of in-context learning in natural language processing research, where large, parameter-heavy models are in use, one pathway that current VICL methods take is model- and data-scaling as key ingredients.
    Yet, it is not clear, whether these ingredients are the key for in-context learning to take shape in vision models.
    To stress-test such large models, we challenge them with an extreme counterexample: we train a tiny visual in-context model with merely $1$ million parameters and a modest amount of $70,000$ images.
    We compare the results of this severely capacity capped tiny model to $7,000\times$ larger VICL models in different adaptive settings, \textcircled{\small{1}} on image data with small distribution shifts, \textcircled{\small{2}} on unseen task encodings and \textcircled{\small{3}} on a completely new task, \ie, the setting VICL envisions.
    With the chasm of training resources between the tiny- and large models, our experiments showcase a lack in how adaptive capabilities are measured, with respect to how tasks are encoded, which tasks were used in pre-training and the choice of metrics.
    These gaps in current VICL benchmarking underscore a need for innovation in evaluation of adaptive capabilities.
\end{abstract}

\thispagestyle{preprint}

\section{Introduction}
In recent years, large pre-trained deep learning models have been validated as tremendously useful backbones for downstream vision tasks~\cite{dosovitskiy2020image,rombach2022high,simeoni2025dinov3,ravi2024sam2}.
With methods from transfer learning~\cite{caruana1994learning,bengio2011deep,yosinski2014transferable}, \eg, fine-tuning large, pre-trained models with a set of labels, a diverse set of vision tasks can be addressed with high performance~\cite{simeoni2025dinov3}.
Yet, the process of annotating image data to obtain a large enough set of labels for fine-tuning can be prohibitively expensive for niche use-cases with small budgets, \eg, in the context of small businesses, scientists without machine learning expertise or use-cases of citizens, such as photography.

Such low budget settings and the expensive process of data annotation~\cite{Cordts2016Cityscapes} has led to research in approaches that require only few labels.
One research area is semi-supervised learning, which requires a few labeled examples and a large amount of unlabeled data~\cite{sohn2020fixmatch,tarvainen2017mean}.
Yet, a large amount of unlabeled data may not be readily available in niche use-cases. 
Another area is few-shot learning~\cite{finn2017model,sung2018learning,ravi2017optimization}, where the prerequisite is merely having access to a small set of labeled data.
Here, the idea resides in using a meta-learning dataset to obtain a model, that quickly adapts to new tasks, although, often, computer vision tasks are considered in isolation, \eg, few-shot classification or few-shot segmentation. 
Recent advances in natural language processing~\cite{brown2020language} show that few-shot examples provided at test time enable large language models to perform  so-called in-context learning. This learning paradigm, which does not require weight updates, has sparked renewed interest in few-shot settings for vision as well.

In ensuing explorations of visual in-context learning~\cite{bar2022visual,bai2024sequential,guo2024data,wang2023promptdiffusion}, different architectures and options for conditioning models on the few-shot examples~\cite{czolbe2023neuralizer,wang2023promptdiffusion,bai2024sequential}, different model scales~\cite{bai2024sequential,guo2024data}, and different training data~\cite{bar2022visual} have been investigated.
In our work, we take a close look at the architectural dimension, particularly at the assumption of scaling model parameters as fundamental component for VICL~\cite{bai2024sequential}.
Therefore, we take a counter intuitive pathway, we investigate how a tiny model compares to up to $7,000\times$ larger models. 
If a severely capacity-capped model can perform competitively, it exposes the possibility that scaling is a brute-force substitute for fundamental innovation in efficient architectures.

In particular, we ask: how does this tiny model perform in adapting to new data distributions, new task encodings and to unseen tasks?
In doing so, we also explore a training recipe for tiny VICL models, including the effect of a new similarity-based loss, the \emph{Palette-aware Dice loss}.
Our contributions amount to:
\begin{itemize}
    \item We train the parameter-efficient \methodname~model on general vision data using a new \emph{Palette-aware Dice loss} function.
    \item We analyze successively larger shifts in terms of image and task distribution for existing visual in-context learning models and our \methodname~model.
    \item With these results, we gather insights with respect to the trend towards large, parameter-heavy models and the evaluation practices in VICL. 
\end{itemize}


\section{Related work}

\subsection{Vision Foundation Models}
In recent years, scaling neural networks~\cite{rombach2022high,dosovitskiy2020image,cherti2023reproducible} and their training data~\cite{schuhmann2022laion} has been a key contributor to pushing the state of the art in computer vision forward.
With sufficient scale, these models can be adapted to a wide range of downstream tasks, leading to the term foundation models~\cite{bommasani2021opportunities}.
Vision foundation models have historically followed empirical neural scaling laws~\cite{zhai2022scaling}, where zero-shot and few-shot classification capabilities improve with exponential increase in model capacity and data volume.
This arms race is largely driven by raw parameter expansion \cite{dehghani2023scaling} paired with scalable self-supervised learning paradigms~\cite{caron2021emerging,he2022masked}, such as DINOv3~\cite{simeoni2025dinov3}, which effectively harness massive amounts of unlabeled images and has led to models learning more expressive image representations.


\subsection{Visual In-Context Learning}
Coinciding with this trend of increasing model sizes, which prominently took place in the natural language processing community~\cite{vaswani2017attention,hestness2017deep,hoffmann2022training,kaplan2020scaling}, it was found, that large language models can be prompted with few examples of a task and a query input, and predict the task-specific output, without weight updates~\cite{brown2020language}.

Following this breakthrough observation of in-context learning in large language models, Visual In-Context Learning (VICL) has emerged as a paradigm for training adaptive vision models that are capable of similar contextual predictions without weight updates~\cite{bar2022visual}.
Specifically, the idea of VICL is to use a dynamic context set, which consists of input-output image pairs that specify the task, \eg, image pairs showing a semantic segmentation, an image colorization, or a depth estimation task.
Alongside this context, a query image is supplied, such that the model acts as a general-purpose inference engine, which derives the underlying task solely from the provided context and applies it to the query.

The earliest VICL approach~\cite{bar2022visual} framed the contextual prediction task as an image inpainting problem, concatenating context and query images into a spatial grid and feed it into a masked autoencoder~\cite{he2022masked} to predict the missing output.
Another line of work~\cite{bai2024sequential,guo2024data} has treated VICL as a sequence modeling task, where input-output image pairs are arranged into \emph{visual sentences}, \ie, sequences of images.
Through visual codebooks~\cite{van2017neural,esser2021taming,rombach2022high}, these image sentences are tokenized, yielding shorter and discrete sequences.
Based on these sequences, the training recipe from language modeling can be mirrored by training large, auto-regressive transformer models~\cite{touvron2023llama} with next-token prediction~\cite{radford2018improving} on large datasets collections.
As such, this approach scales model and dataset size, and computational load in training, to investigate, whether in-context learning properties arise.


In contrast, the medical domain has to cope with data scarcity~\cite{reiss2021every,seibold2022reference,reiss2022graph,reiss2023decoupled,yu2019uncertainty}, which led to the development of efficient in-context learning models on much smaller scales~\cite{czolbe2023neuralizer,butoi2023universeg,rakic2024tyche}.
In particular, the Neuralizer architecture~\cite{czolbe2023neuralizer} follows a design with its pairwise conv-average blocks, which efficiently enables processing contextual images aside a query image.
While the architecture has been trained for different medical domains~\cite{czolbe2023neuralizer,negrini2025conquering}, it has not been tested on general vision tasks.
In this work, we close this gap and train a $1$M parameter model on common vision tasks to investigate to what extent a tiny model can approach the performance of models with up to $7$B parameters.




\section{Methodology}
Next, we introduce the VICL paradigm, the training recipe for our~\methodname~model, including training datasets and the \emph{Palette-aware Dice loss} function. 

\subsection{Preliminaries}
\subsubsection{Visual in-Context Learning}
In visual in-context learning, a model $\theta$ is designed to, given a context set $\mathcal{C} = \{(c_{in}^0,c_{out}^0), \dots, (c_{in}^N,c_{out}^N)\} \in \mathbb{R}^{N \times 2 \times 3 \times W \times H}$,
in the form of $N$ few-shot image-pairs and a query image $q \in \mathbb{R}^{3 \times W \times H}$, predict a respective output $y_{pred} = \theta(\mathcal{C}, q) \in \mathbb{R}^{3 \times W \times H}$.
Here, each few-shot example refers to an input-output transformation, \ie, the task definition, which may be a segmentation task ($c_{in}$ RGB image, $c_{out}$ segmentation), image colorization ($c_{in}$ grayscale image, $c_{out}$ RGB image), depth estimation  ($c_{in}$ RGB image, $c_{out}$ depth map), or any other dense prediction task. 
In summary, the idea of VICL is to solve the task for the query image $q$, which is defined in the context set $\mathcal{C}$.
Such a contextual prediction is especially interesting for context sets which define tasks that were not used for training the model $\theta(\cdot, \cdot)$.
Training such contextual vision models is most commonly done on a set of diverse vision tasks.

\subsubsection{Neuralizer Architecture}

In contrast to a trend of scaling VICL model size~\cite{bai2024sequential,guo2024data}, the Neuralizer has been proposed in the medical image processing community for efficient contextual prediction~\cite{czolbe2023neuralizer}.
It is based on a simple Unet~\cite{ronneberger2015u}, and successively fuses information of the context set $\mathcal{C}$ and the query image $q$ into intermediate features at different hierarchies through the pairwise conv-average block.
The Retinalizer model~\cite{negrini2025conquering} slightly adapted Neuralizer to predicting three-channel outputs, with the total architecture counting merely $1$M parameters.
We take this variant and train the \methodname~model on general vision tasks to match it up against billion parameter VICL models. 

\subsection{Training a Tiny Visual in-Context Learner}

Next, we discuss how we train the~\methodname~model, describing the multi-task pre-training datasets, training strategy and loss functions.
As multi-task training data, we use the following diverse set of eight datasets with seven training tasks.

\begin{figure}[b]
    \centering
    \rotatebox{90}{$\phantom{.......}$ $c_{out}$ $\phantom{............}$ $c_{in}$} \hfill \includegraphics[width=0.975\textwidth, trim={0 2.1cm 0 0}, clip]{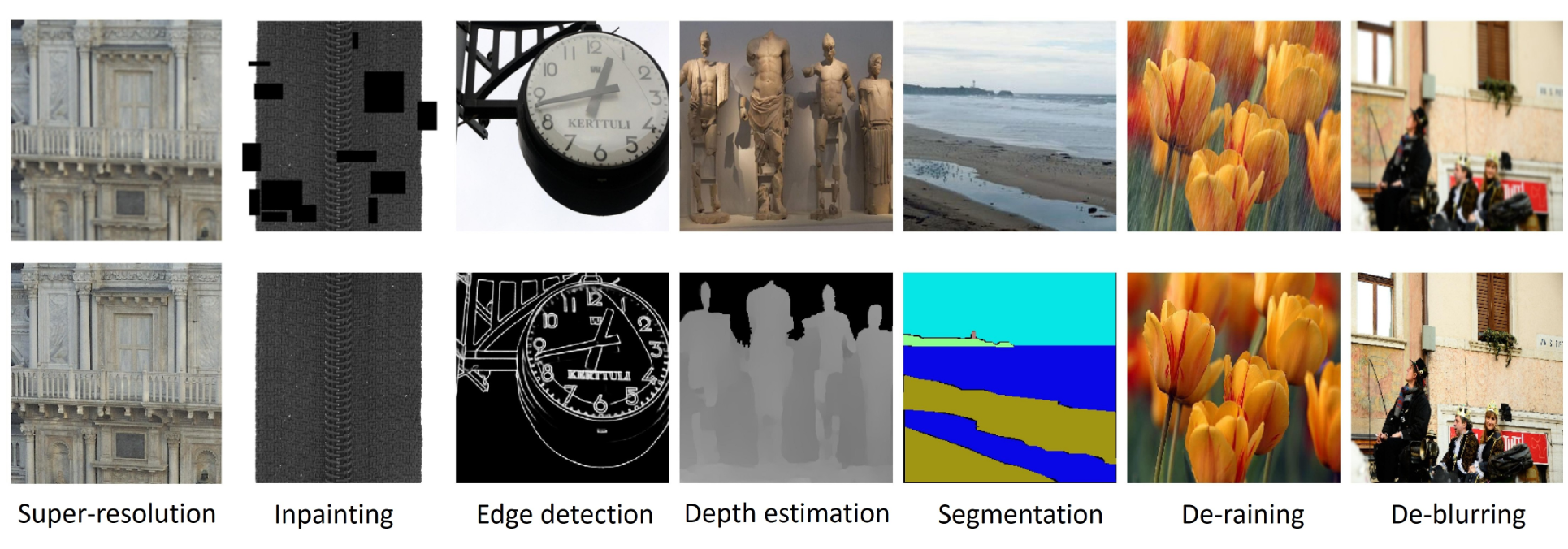}
    \tiny
    \begin{tabular}{c p{1.68cm}p{1.68cm}p{1.68cm}p{1.68cm}p{1.68cm}p{1.68cm}p{1.68cm}}
         $\phantom{...}$& Super-res. & Inpainting & Edge det. & Depth est. & Semantic Seg. & De-rain. & Denoising\\
    \end{tabular}
    \caption{Examples images of the task data used in multi-task pre-training, first row shows the inputs, second row shows the outputs for the different tasks in the columns.}
    \label{fig:tasks_datasets}
\end{figure}

\noindent\textbf{Rain13k~\cite{jiang2020multi} -- De-raining} We directly use the dataset with the provided image annotations, it is a dataset of $13,711$ paired clear and rainy images, the task resides in obtaining a clear version from the rainy image.

\noindent\textbf{HRWSI~\cite{xian2020structure}  -- Depth estimation} As dense regression task in the set of pre-training tasks, we use depth estimation.
Therefore, we use the High-Resolution Web Stereo Image dataset with $20,378$ images paired with depth maps.

\noindent\textbf{ADE20K~\cite{zhou2017scene}  -- Semantic segmentation} As pixel-wise classification task, we use ADE20K, which covers a broad set of $150$ classes in $20,210$ images.

\noindent\textbf{MVTec AD~\cite{bergmann2021mvtec} -- Inpainting} To broaden the image distribution beyond images from everyday life, we add the MVTec AD dataset, which contains different textures and object categories.
We use the $3,629$ images for the inpainting task.

\noindent\textbf{RAISE~\cite{dang2015raise} -- Denoising, Super-resolution} We use the $2,000$ high resolution images in this dataset for the tasks of image denoising, \ie, removing Gaussian noise, and recovering high-resolution details from down-scaled images.

\noindent\textbf{BIPED~\cite{poma2020dense}, MDBD~\cite{mely2016systematic}, COCO~\cite{lin2014microsoft} -- Edge detection} For edge detection, the dataset is created by collecting $7,500$ images from BIPED, MDBD and a subset of COCO with annotations created with the protocol of DexiNed~\cite{poma2020dense}.

In~\Cref{fig:tasks_datasets}, we depict images and task outputs from the respective datasets.
For training, we split the data into $60\%$ train / $20\%$ val / $20\%$ test partitions.

\paragraph{Task sampling}
As different datasets come with wildly different numbers of images, uniform batch sampling heavily skews the training towards the larger datasets and their tasks.
To prevent biased learning, a weighted sampling technique balances the contribution across tasks~\cite{negrini2025conquering,reiss2025visual}.
Let $D_t$ denote the dataset associated with task $t$, where $t \in (1, ...., T)$ indexes the set of available tasks and $|D_t|$ denotes the number of samples within the dataset.
Each sample $i \in D_t$ is assigned the initial weight $w_i = \frac{1}{|D_t|}$. 
Let $\mathcal{I}$ be the set of all samples drawn from all tasks, indexed by $j$.
The normalized sampling probability for sample $i$ is defined as ${w^\star_i} = \frac{w_i}{\sum_{j\in \mathcal{I}} w_j}$, where the denominator ensures that $\sum_{j\in \mathcal{I}} w_j = 1$.
At each training iteration, sample indices are drawn independently according to the multinomial distribution induced by the probability mass function $\{w^\star_i\}_{i \in \mathcal{I}}$.

\paragraph{Loss functions}
For training the \methodname~model, we use a set of loss functions, which incentivize different aspects for different vision tasks.
The most general, standard loss is the MSE loss:
\begin{equation}
    \mathcal{L}_{MSE} = ||y-y_{pred}||^2\enspace,
    \label{eq:mse}
\end{equation}
with the prediction $y_{pred}$ and the target $y$, applied in element-wise fashion, thereby, penalizing fine-granular deviations.
Further, we experiment with the DSSIM loss, a loss function based on the Structural Similarity (SSIM)~\cite{wang2004image} score:
\begin{equation}
    \mathcal{L}_{DSSIM} = 1 - SSIM(y, y_{pred})\enspace.
    \label{eq:dssim}
\end{equation}
Finally, we put forward a special loss function aimed at color coherent prediction for VICL, which we term the \emph{Palette-aware Dice loss (PaD)}.
For a given target image $y$, the set of unique normalized RGB colors $\mathcal{U} = \{c_0, \dots, c_k\}, c_i \in [0,1]^3$ is extracted.
With this color palette, we compute $k$ binary maps:
\begin{equation}
    \hat{y}_j^k = \frac{y_j^Tc_k}{\lVert y_j\rVert~\lVert c_k\rVert} \doteq 1\enspace,
\end{equation}
which puts ones at spatial positions containing the color $c_k$ and zeros elsewhere.
Then, based on a similarity measure $\sigma(\cdot, \cdot)$, a softmax color match is computed:
\begin{equation}
    s_j^i = \frac{\text{exp}(\sigma(c_i, y_{pred, j}))}{\sum_{l\in [0, \dots, k]} \text{exp}(\sigma(c_l, y_{pred, j}))}\enspace.
    \label{eq:colormatchscore}
\end{equation}
Based on these scores and the Dice loss~\cite{sudre2017generalised}, the \emph{PaD loss} function can be set up:
\begin{equation}
    \mathcal{L}_{PaD}^i = \frac{2 \sum_{j \in \Omega} s^i_j \hat{y}^i_j}{\varepsilon + \sum_{j \in \Omega} s_j^i + \sum_{j \in \Omega} \hat{y}^i_j}\enspace,
\end{equation}
with $\Omega$ denoting the set of pixel indices. The final loss function is averaged across all colors in $\mathcal{U}$:
\begin{equation}
    \mathcal{L}_{PaD} = 1 - \frac{1}{k + 1} \sum_{l \in [0, \dots, k]} \mathcal{L}_{PaD}^l\enspace.
\end{equation}
This formulation enforces color fidelity and encourages predicted RGB values to align with predefined class colors in the context set, which is especially important for tasks like semantic segmentation.
The similarity function in~\Cref{eq:colormatchscore}, can be chosen differently. We use Euclidean-based similarity $\sigma(a,b) = - \lVert a-b\rVert_2$. 

\paragraph{Implementation Details}
All training runs are executed on a GeForce RTX 2080 Ti GPU. We ablate the loss functions on a Unet-based Neuralizer architecture instantiated with 4 encoder and decoder layers each containing $64$ channels leading to $1$M parameters.
We optimized the parameters with ADAM~\cite{kingma2014adam} and a learning rate of $0.0001$ with early stopping after ten epochs.
Training was done with batch size three and seven context pairs. All images were of size $256 \times 256$.

\section{Evaluation}

\subsection{In-Domain Prompting}
First, we are interested in prompting our trained \methodname~model with tasks that have been seen during training. However, the evaluation is performed on test images that were not included in training, \ie, a standard multi-task evaluation.

\begin{table}[b]
    \caption{Multi-task quantitative evaluation results for the \methodname~model with different loss functions (best scores in bold, $\downarrow$ lower is better and $\uparrow$ higher is better).}
    \label{tab:multi_task_results}
    \centering
    \resizebox{\textwidth}{!}{
    \begin{tabular}{l  c c  c c  c c  c c  c c  c c c  c}
        \toprule
        \multirow{2}{*}{\textbf{Model\_id}}
        & \multicolumn{2}{c}{\textbf{Depth Est.}} 
        & \multicolumn{2}{c}{\textbf{Denoising}} 
        & \multicolumn{2}{c}{\textbf{Inpainting}} 
        & \multicolumn{2}{c}{\textbf{De-raining}} 
        & \multicolumn{2}{c}{\textbf{Super-res.}} 
        & \multicolumn{3}{c}{\textbf{Edge Detection}} 
        & \textbf{Sem. Seg.} \\
        & RMSE $\downarrow$ & MAE $\downarrow$ 
        & RMSE $\downarrow$ & MAE $\downarrow$ 
        & RMSE $\downarrow$ & MAE $\downarrow$ 
        & RMSE $\downarrow$ & MAE $\downarrow$ 
        & RMSE $\downarrow$ & MAE $\downarrow$ 
        & Prec. $\uparrow$& Recall $\uparrow$& F1 $\uparrow$ 
        & IoU $\uparrow$ \\
        \midrule
        $\mathcal{L}_{MSE}$ & 0.18186 & 0.14808 & 0.06880 & 0.04099 & 0.03040 & 0.01135 & 0.04509 & 0.03209 & \textbf{0.06457} & 0.03607 & \textbf{0.81455} & 0.70946 & 0.75505 & 0.01463 \\
        $^\llcorner+\mathcal{L}_{DSSIM}$ & \textbf{0.17830} & \textbf{0.14453} & \textbf{0.06841} & \textbf{0.04076} & \textbf{0.02512} & \textbf{0.00759} & 0.03820 & 0.02724 & 0.06601 & \textbf{0.03576} & 0.81276 & \textbf{0.74441} & \textbf{0.77497} & 0.01521 \\
        $~~^\llcorner+\mathcal{L}_{PaD}$ & 0.18281 & 0.14942 & 0.06860 & 0.04099 & 0.02594 & 0.00797 & \textbf{0.03795} & \textbf{0.02712} & 0.06623 & 0.03588 & 0.80866 & 0.73907 & 0.76998 & \textbf{0.01777} \\
        \bottomrule
    \end{tabular}
    }
\end{table}

\subsubsection{Quantitative results}
First, we look into quantitative results of the 1M parameter Unet variant in~\Cref{tab:multi_task_results}.
Training the model with $\mathcal{L}_{MSE}$ shows a base performance, indicating that even a small model is able to fit the variety of tasks to some extent.
Although, what is apparent are the low performance metrics on the segmentation task.
This is partly due to the $150$ classes to be segmented, but also due to the ill-fitting loss function producing blurry segmentation predictions, which may be due to $\mathcal{L}_{MSE}$ struggling with multi-modal color distributions, \eg, as seen previously in image colorization~\cite{larsson2016learning}. 
Adding the $\mathcal{L}_{DSSIM}$ loss leads to improvements throughout, as it helps to preserve fine structures. This is essential for many of the tasks, \eg, restoration tasks such as de-blurring, inpainting, de-raining, super-resolution, but also more semantic tasks like edge detection or segmentation.
Finally, we add the $\mathcal{L}_{PaD}$ with Euclidean similarity scoring $\sigma_e$ in row three of~\Cref{tab:multi_task_results}.
While the metrics are slightly lower for the majority of tasks, the added loss substantially increases the segmentation performance of the~\methodname~model by $16.8\%$.
Our $\mathcal{L}_{PaD}$ clearly bridges the gap between regression-based losses like $\mathcal{L}_{MSE}$ and the discrete nature of semantic segmentation classes. Hence, for the subsequent evaluations of out-of-domain prompting, we explore the 1M variant, trained with $\mathcal{L}_{MSE} + \mathcal{L}_{DSSIM} + \mathcal{L}_{PaD}$.

\subsubsection{Qualitative results}
\begin{figure}[t]
    \centering
    \includegraphics[width=0.98\textwidth,page=1]{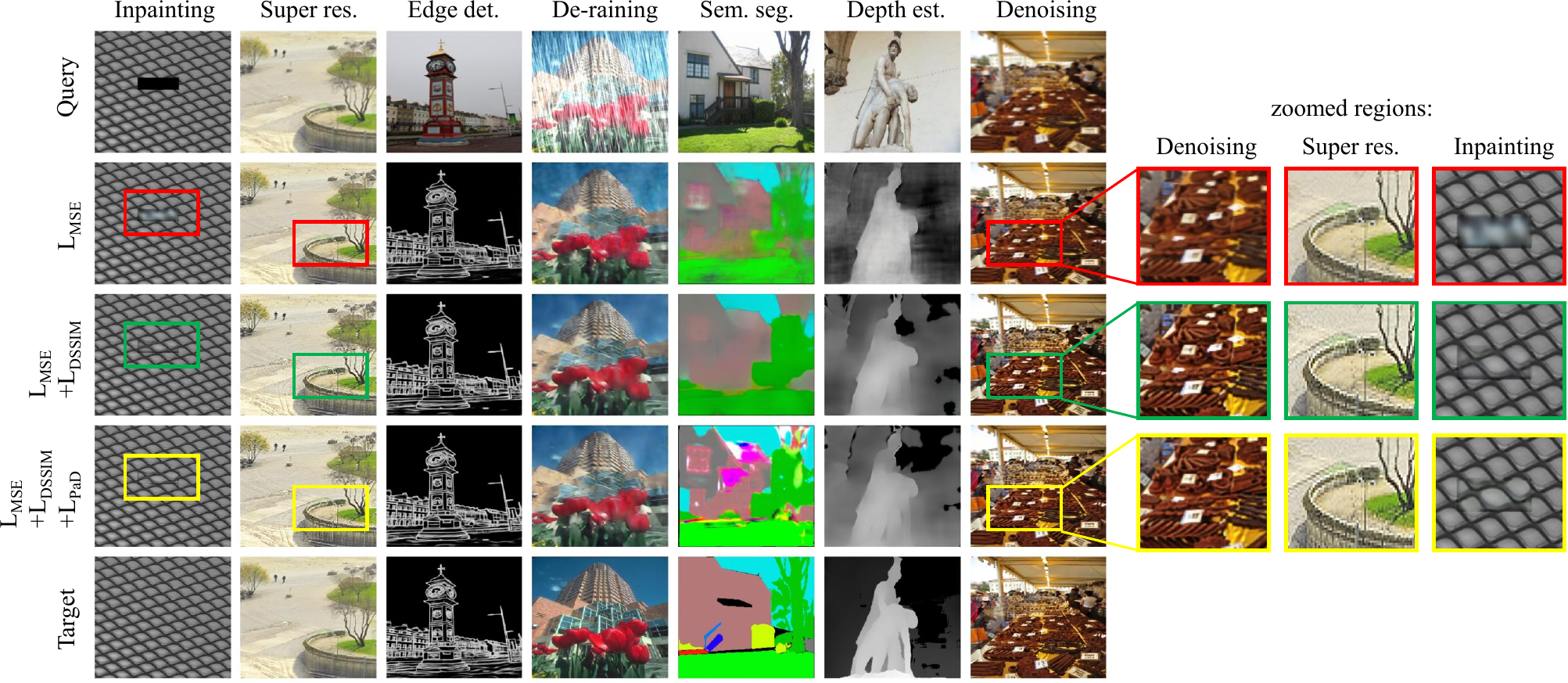}
    \caption{Qualitative results for \methodname~with different losses. While results are imperfect, the low capacity model with $1$M parameters learns to address the seven tasks.}
    \label{fig:multitask_qualitative}
\end{figure}

In~\Cref{fig:multitask_qualitative}, we see the qualitative results of \methodname.
While far from perfect, the parameter-constrained model clearly captures the core aspects of all pre-training tasks, performing better on some (\eg, edge detection and denoising) than on others (\eg, semantic segmentation).
With these clear capabilities and limitations, we can ask, how the training recipe of VICL models, \ie, training in a multi-task, image-to-image setup, transfers to tiny models and how it compares to models with up to $7,000\times$ more parameters.


\subsection{Out-of-domain Prompting}
\label{sec:oodprompting}
In Visual in-Context Learning, prompting models with either new image distributions or new task distributions is particularly interesting.
To explore this, we successively shift away from the pre-training image and task distribution, by \textcircled{\small{1}} prompting our \methodname~model with tasks it was trained on but using images from a different image distribution, \textcircled{\small{2}} prompting the model with tasks it was trained on but using a different task encoding than was used during training, and \textcircled{\small{3}} prompting the model with previously unseen tasks, \ie, tasks that are out of its pre-training task distribution.

Specifically, we take a look at the following datasets and tasks:

\noindent\textbf{DID-MDN -- De-raining} This dataset is a de-raining dataset, \ie, a seen task for our \methodname~model.
It encompasses a new image distribution not directly used in training, marking a small shift.

\noindent\textbf{BSDS500 -- Edge Detection} This dataset slightly shifts the image distribution and contains labels for the task edge detection seen in training by \methodname.

\noindent\textbf{PASCAL VOC -- Single Object Segmentation} This segmentation task is different from the pre-training segmentation tasks, as it encodes the task as segmenting prominent foreground objects in white and the background in black. 
This is a shift in the task encoding for~\methodname, as it was trained on segmentation with a fixed color map and on scene segmentation rather than object segmentation.

\noindent\textbf{NYU Depth V2 -- Depth Estimation} \methodname~was trained on depth estimation, yet, it was trained on grayscale maps, \ie, near $\to$ far as light $\to$ dark.
This task is encoded differently by using a blue $\to$ red color map.

\noindent\textbf{ImageNet -- Image Colorization} While the image distribution is a small shift to pre-training tasks, \methodname~was not trained on the image colorization task.

Next, we describe the VICL models that we compare our \methodname~to.

\noindent \textbf{MAE-VQGAN}~\cite{bar2022visual} is a VICL model trained on uncurated figures from computer vision papers, it makes use of a masked image modeling objective~\cite{he2022masked} on top of a visual codebook~\cite{esser2021taming}, the trained backbone is a Vision Transformer~\cite{dosovitskiy2020image}.

\noindent \textbf{LVM}~\cite{bai2024sequential} is a large transformer model with $7$B parameters based on the Llama architecture~\cite{touvron2023llama} and forms visual sentences, which are tokenized with a pre-trained codebook~\cite{esser2021taming}.
It is trained on broad vision tasks from $50$ image datasets.

\noindent \textbf{DeLVM}~\cite{guo2024data} is similar to LVM, but considers a smaller set of pre-training tasks and trains a $1$B parameter model and distills a $300$M parameter model from it.

\noindent \textbf{Copy-query} is a naive baseline, which takes the query image as prediction.

\noindent \textbf{Copy-target} is similar, but returns a random target image from the context.

All models are prompted with context size $|C| = 2$, \ie, two context pairs, only MAE-VQGAN is prompted with $|C| = 1$ due to its image grid structure.

\subsubsection{Quantitative results}
\subsubsection{Training-consistent Task Prompts}
First, we consider the setting \textcircled{\small{1}}, \ie, tasks encoded in the prompt are consistent with pre-training tasks of~\methodname, but the dataset, \ie, the data distribution is different.
Specifically, we consider edge detection and de-raining in~\Cref{tab:deraining_edge_multitask}.
Accross all metrics, the \methodname~model consistently outperforms the competing approaches, indicating strong robustness to slight shifts in data distribution.
Further, opposing models, adapt less well to the task prompt. However, it should be noted that LVM and MAE-VQGAN were not explicitly trained on the deraining task, which likely contributes to their comparatively lower performance.
Results on the DeLVM task are also worse, even though, it has been trained on de-raining and uses $\times 1,000$ more parameters.
A possible explanation is DeLVM's reliance on a visual codebook, which limits the granularity of the image reconstruction.
Because the quantization error from the codebook degrades pixel-wise metric scores (SSIM, PSNR, MSE), we argue that a codebook-free VICL baseline is essential for an accurate assessment.

\begin{table}[t]
\centering
\small
\caption{Evaluation results for de-raining on DID-MDN and edge detection on BSDS500 (best metric scores in bold, $\downarrow$ lower is better and $\uparrow$ higher is better).}
\label{tab:deraining_edge_multitask}
\begin{tabular}{l c c c c  c c c c}
\toprule
\multirow{2}{*}{\textbf{Model}} & \multicolumn{4}{c}{\textbf{De-raining}} & $\phantom{-}$& \multicolumn{3}{c}{\textbf{Edge Detection}} \\
& \scriptsize LPIPS$\downarrow$ & \scriptsize SSIM$\uparrow$ & \scriptsize PSNR$\uparrow$ & \scriptsize MSE$\downarrow$ & & \scriptsize Prec. $\uparrow$ & \scriptsize Recall$\uparrow$ & \scriptsize F1$\uparrow$ \\
\midrule
Copy Query & 0.2744 & 0.6786 & 19.727 & 0.0161 & & 0.0649 & 0.3184 & 0.0962\\
Copy Context & 0.7620 & 0.1215 & 10.019 & 0.1360 & & 0.0845 & 0.0884 & 0.0782 \\
\midrule
MAE-VQGAN & 0.5434 & 0.2991 & 15.543 & 0.0302 & & 0.1201 & 0.0270 & 0.0397 \\
DeLVM (300M) & 0.1673 & 0.5271 & 21.236 & 0.0094 & & 0.0065 & 0.0009 & 0.0014 \\
DeLVM (1B) & 0.1579 & 0.5389 & 21.409 & 0.0091 & & 0.0068 & 0.0067 & 0.0028 \\
LVM & 0.2640 & 0.4571 & 19.398 & 0.0142 &  & 0.0729 & 0.2689 & 0.0752 \\
\methodname & \textbf{0.1247} & \textbf{0.8476} & \textbf{26.509} & \textbf{0.0031} & & \textbf{0.2587} & \textbf{0.4685} & \textbf{0.3100} \\
\bottomrule
\end{tabular}

\end{table}

While F1 scores for edge detection are far below the task from pre-training, the \methodname~model still performs best, while MAE-VQGAN and DeLVM produce very low scores.
Even LVM, which has been pre-trained on edge detection, fails to address the task.
Upon closer investigation, we see that LVM's edge detection task encoding is different from how we prompt the task.
Specifically, LVM was trained on white background and black edges.
With this adapted encoding, LVM's results increase from $7.29\%$ Precision, $26.89\%$ Recall and $7.52\%$ F1 Score as in~\Cref{tab:deraining_edge_multitask} to $19.72\%$ Precision, $36.19 \%$ Recall and $22.96\%$ F1 Score.
This indicates a failure to adapt to different task encodings, calling into question whether even large models can handle such simple shifts.


\begin{table}[b]
    \centering
    \begin{minipage}{.6\linewidth}
    \caption{Evaluation for segmentation on Pascal  (split 0 - 3) and depth estimation on NYU Depth V2 (best in bold, $\downarrow$ lower better, $\uparrow$ higher better).}
    \label{tab:pascal_depth_multitask}
    \begin{tabular}{l c c c c  c}
        \toprule
        \multirow{2}{*}{\textbf{Model}} & \multicolumn{4}{c}{\textbf{Sem. Seg.} (mIoU $\uparrow$)} & \textbf{Depth} \\
        & sp0 & sp1 & sp2 & sp3 & \scriptsize RMSE$\downarrow$ \\
        \midrule
        Copy Query & 0.08 & 0.13 & 0.10 & 0.19 & 0.5237 \\
        Copy Context & 12.74 & 17.71 & 14.16 & 15.14 & 0.5309 \\
        \midrule
        MAE-VQGAN & \textbf{34.29} & 37.15 & 32.28 & 28.54 & \textbf{0.5012} \\
        DeLVM (300M) & 8.39 & 12.25 & 10.70 & 10.08 & 0.5409 \\
        DeLVM (1B) & 8.45 & 13.20 & 13.86 & 12.54 & 0.5327 \\
        LVM & 32.80 & \textbf{37.33} & \textbf{36.99} & \textbf{30.00} & 0.5375 \\
        \methodname& 13.19 & 16.69 & 12.57 & 13.50 & 0.5526 \\
        \bottomrule
    \end{tabular} \hfill \end{minipage}
    \begin{minipage}{.35\linewidth}
        \centering
        \small
        \caption{Impact of task encoding for depth estimation in RMSE for LVM, and the $7,000\times$ smaller \methodname.
        }
        \label{tab:depth_label_representation}
        \begin{tabular}{l c c}
            \toprule
            \tiny Task encoding & LVM & \methodname \\
            \midrule
            \tiny blue $\to$ red & 0.5375 & 0.5526 \\
            \tiny light $\to$ dark & 0.4675 & 0.3109 \\
            \tiny dark $\to$ light & 0.3321 & \textbf{0.2599} \\
            \bottomrule
        \end{tabular}
    \end{minipage}
\end{table}

\subsubsection{Training-inconsistent Task Prompts}
With the insight that slightly altered edge maps in the prompt already lead to substantially different predictions, we continue to explore this dependency on the task encoding of VICL models further (setting \textcircled{\small{2}}). 
To this end, we prompt \methodname~with segmentation maps that use color encodings different from those used during training, and represent depth maps with a color depth encoding rather than the grayscale map used during training. The results are reported in~\Cref{tab:pascal_depth_multitask}.
Contextual, adaptive models should still produce coherent predictions even under altered task encodings, guided by the example pairs in $\mathcal{C}$.
We see that the \methodname~model is not able to adapt to the differently defined task prompts, falling short of the other VICL models as well as the copy baselines.
Similarly, the DeLVM models fall short, for most segmentation splits even under performing the \methodname~model, with $\times 1,000$ less parameters.
LVM and MAE-VQGAN perform better than the baselines in segmentation. However, LVM has been exposed to a training task very similar to binary segmentation, which is depth estimation with gray value maps that show near structures in white, and far away structures in black.
This task encoding closely reflects binary segmentation on PASCAL VOC with prominent foreground objects to segment, making the need for adaptation obsolete. 

\begin{wraptable}{r}{0.57\textwidth}
    \centering
    \small
    \caption{Evaluation results for image colorization on a subset of ImageNet (best metric scores in bold, $\downarrow$ lower is better and $\uparrow$ higher is better).}
    \label{tab:colorization_multitask}
    \begin{tabular}{l c c c c c}
        \toprule
        \multirow{2}{*}{\textbf{Model}} & \multicolumn{4}{c}{\textbf{Colorization}} \\
        & \scriptsize LPIPS$\downarrow$ & \scriptsize SSIM$\uparrow$ & \scriptsize PSNR$\uparrow$ & \scriptsize MSE$\downarrow$ \\
        \midrule
        Copy Query & 0.2193 & 0.9143 & 23.817 & 0.0089 \\
        Copy Context & 0.7563 & 0.1211 & 8.6545 & 0.1505 \\
        \midrule
        \methodname & \textbf{0.2672} & \textbf{0.8814} & \textbf{22.520} & \textbf{0.0098} \\
        LVM & 0.2904 & 0.5294 & 18.448 & 0.0182 \\
        MAE-VQGAN & 0.5623 & 0.3535 & 15.192 & 0.0347 \\
        DeLVM (300M) & 0.5445 & 0.3727 & 14.311 & 0.1052 \\
        DeLVM (1B) & 0.3189 & 0.5727 & 18.976 & 0.0203 \\
        \bottomrule
    \end{tabular}
\end{wraptable}
On the other hand, when considering depth estimation with color encoding, all models apart from MAE-VQGAN fail to adapt to the task encoding and perform worse than the naive copy baselines, irrespective of model size.
Investigating the extremes, \ie, LVM and \methodname~more precisely, by adapting the color encoding for depth estimation to match the training setups of the two models, we see results shifting in~\Cref{tab:depth_label_representation}.
Moving away from the blue $\to$ red encoding, towards a grayscale encoding with close structures encoded bright and far away structures dark in row two leads to substantial improvements.
Matching the task encoding as in the training set of \methodname, as in row three of~\Cref{tab:depth_label_representation}, we see the best depth estimation results, with \methodname's RSME more than halved.

These results show that model size alone does not lead to improved adaptation. Furthermore, training on more data in itself is not the key either, but merely increases the chance of a training task aligning with a new task.

\subsubsection{Unseen Task Prompts}
Finally, we prompt with a task not present in the training of \methodname~(setting \textcircled{\small{3}}), the image colorization task in~\Cref{tab:colorization_multitask}.
Even though, LVM has been trained on image colorization, we see that the \methodname~model outperforms all VICL models with lowest LPIPS and MSE values and with the highest SSIM and PSNR values.
We also see that none of the trained models is able to outperform the naive baseline of simply copying the grayscale image as the prediction.
On first glance, this seems to indicate that none of the models is able to address the task. Among the VICL models, \methodname~comes closest to correctly addressing the task of image colorization.
However, this is not the case, as we see in the qualitative results next.

\subsubsection{Qualitative results}
\begin{figure}[t]
    $\phantom{........}$ Query $\phantom{......}$ \methodname $\phantom{.....}$ MAE-VQG $\phantom{......}$ LVM $\phantom{.........}$ DeLVM $\phantom{........}$ Target
    
    \rotatebox{90}{\small {\tiny edge det.} \textcircled{\tiny{1}}$\phantom{...}${\tiny depth est.} \textcircled{\tiny{2}}$\phantom{..}${\tiny binary seg.} \textcircled{\tiny{2}}$\phantom{...}${\tiny de-rain.} \textcircled{\tiny{1}}$\phantom{....}${\tiny colorize} \textcircled{\tiny{3}}}\hfill\includegraphics[width=0.98\textwidth]{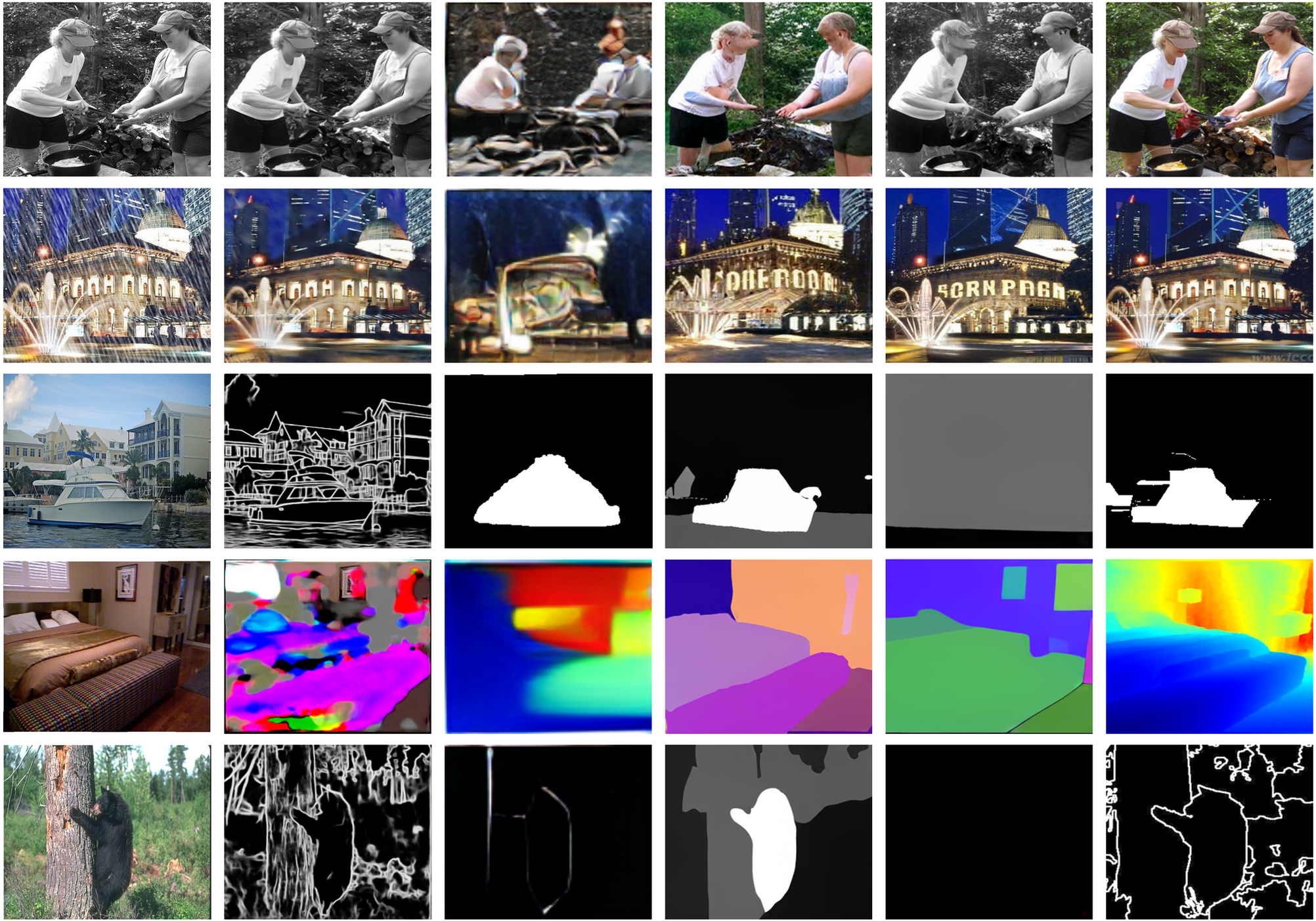}
    \caption{Qualitative results for all VICL models for out-of-domain prompting.}
    \label{fig:ood_qualitative}
\end{figure}

In~\Cref{fig:ood_qualitative}, we show the qualitative results for prompting the models with the different out-of-domain prompts (settings \textcircled{\small{1}}, \textcircled{\small{2}}, \textcircled{\small{3}}).
Starting with colorization in row one, we can see that the qualitative results are not coherent with a straightforward interpretation of the quantitative results, namely, that copy query and \methodname~yield the best results.
LVM for example, which was trained on colorization, is able to coherently colorize the image.
Yet, what leads to lower metric scores is, that pixel-wise handing through the input, as, \eg, \methodname~does, produces a lower error. 
This is in contrast to LVM which uses a visual codebook and thereby is not able to reproduce fine-grained structures from query, yielding a worse score, but actually colorizes the image.

Similarly, for de-raining, all models are able to predict images without the rainy pattern, yet, MAE-VQGAN, LVM, and DeLVM all alter the image content severely due to their use of codebooks.
DeLVM even hallucinates illuminated letters in place of windows.
For most models, depth estimation with colorized depth maps triggers segmentation, only MAE-VQGAN follows the prompt coarsely. 

Edge detection is a task from training for \methodname, thus, the prediction is fairly accurate. Yet, as it was pre-trained on more fine-grained edge maps, it detects more details irrespective of the edge task in the prompt.
Finally, prompting with the binary segmentation of a ship leads to good results for MAE-VQGAN and LVM, while DeLVM fails and \methodname~detects edges, hinting at superficial patterns that were learned to determine which pre-training task to infer.

\section{Discussion and Conclusion}
In this work, we showcased the behavior of different models in three settings: \textcircled{\small{1}} changing the image data distribution of a seen task \textcircled{\small{2}} changing the encoding of a seen task and \textcircled{\small{3}} changing to an unseen task.
Next, we discuss the insights when comparing the \methodname~model with models of $7,000\times$ its size.

\paragraph{Visual in-Context Learners can bridge small domain gaps}
When a model is trained on a task, and addresses the task well on an in-domain data, the resulting model can address the same task on new data, if the domain shift is small as seen in setting \textcircled{\small{1}}.
Yet, as soon as the data distribution, or the task encoding shifts too much, models are incapable to adapt beyond learned tasks as in setting \textcircled{\small{2}}.

\paragraph{Visual in-Context Learners do not predict contextually}
From the results in~\Cref{fig:ood_qualitative}, we observe that when context-prompts deviate substantially from seen tasks, models still tend to revert to behaviors aligned with these pre-training tasks. For example, \methodname~fails to adapt to the sparser edges, while LVM and DeLVM produce segmentations consistent with their pre-training, rather than the intended color depth maps.
This behavior suggests that the models do not fully perform contextual inference, but instead rely on a superficial routing towards one of the learned output modalities based on the provided context.

\paragraph{Traditional metrics do not directly reflect good contextual prediction}
Prompting VICL models with the colorization task reveals a discrepancy between quantitative and qualitative results. This highlights that conventional metrics fail to capture the essence of contextual prediction that is central to VICL.
Rather than rewarding the appropriate colorization by LVM, the metrics penalize pixel-wise inaccuracies, and even the perceptual metric LPIPS is not equipped to attribute better scores to the models that actually address the task. 
This effect is even more pronounced in architectures that allow direct propagation of pixel values and can therefore learn the identity function. For example, \methodname, due to its skip-connections and direct RGB value regression, attains better metric scores than approaches based on discrete codebooks. \\


\noindent Ultimately, our findings reveal that simply scaling model size does not overcome the bottleneck of VICL: reliance on superficial task routing rather than true contextual prediction.
To advance, research should shift from parameter (and dataset) expansion toward designing context-sensitive metrics and architectural- or data-distributional~\cite{chan2022data,bratulic2025unlocking} methods for ensuring context-anchored predictions. \\

\noindent\textbf{Acknowledgements}
This work was supported by funding from the pilot program Core-Informatics of the Helmholtz Association (HGF). The authors acknowledge support by the state of Baden-Württemberg through bwHPC. This work was performed with the help of the Large Scale Data Facility at the Karlsruhe Institute of Technology funded by the Ministry of Science, Research and the Arts Baden-Württemberg and by the Federal Ministry of Education and Research.

\bibliographystyle{splncs04}
\bibliography{egbib}

\end{document}